%% file: main.tex
\documentclass[a4paper,11pt]{article}

\usepackage[utf8]{inputenc}
\usepackage[T1]{fontenc}
\usepackage{times}      
\usepackage{xspace}     
\usepackage{subcaption}

\usepackage{indentfirst}    
\usepackage{amsmath,amssymb,amsfonts}
\usepackage{bm}         

\usepackage[margin=1in]{geometry} 
\usepackage{setspace}             

\usepackage{graphicx}
\usepackage{xcolor}
\usepackage{booktabs}   
\usepackage{multirow}   
\usepackage{tcolorbox}  

\usepackage{enumitem}   
\usepackage{algorithm}
\usepackage{algorithmic}

\usepackage[numbers,square,sort&compress]{natbib}

\usepackage[normalem]{ulem}

\usepackage{url}
\usepackage{hyperref}
\usepackage{cleveref}   

\newcommand{\method}{\textsc{InSpatio-World}\xspace}
\newcommand{\E}{\mathbb{E}}

\newcommand{\Loss}{\mathcal{L}}
\newcommand{\x}{\bm{x}}

\definecolor{bestcolor}{HTML}{D4EDDA}

\title{\method: A Real-Time 4D World Simulator via Spatiotemporal Autoregressive Modeling}

\author{
  \textbf{InSpatio Team (Alphabetical Order):} \\
  \textbf{D}onghui Shen, \textbf{G}uofeng Zhang, \textbf{H}aomin Liu, \textbf{H}aoyu Ji, \textbf{H}ujun Bao, \textbf{H}ongjia Zhai, \\
  \textbf{J}ialin Liu, \textbf{J}ing Guo, \textbf{N}an Wang, \textbf{S}iji Pan, \textbf{W}eihong Pan, \textbf{W}eijian Xie,\\
   \textbf{X}ianbin Liu, \textbf{X}iaojun Xiang,\textbf{X}iaoyu Zhang, \textbf{X}inyu Chen, \textbf{Y}ifu Wang, \\
  \textbf{Y}ipeng Chen, \textbf{Z}henzhou Fan, \textbf{Z}hewen Le, \textbf{Z}hichao Ye, \textbf{Z}iqiang Zhao.
}

\date{}

\begin{document}
\maketitle

\begin{figure}[!ht]
    \centering
    \includegraphics[width=\textwidth]{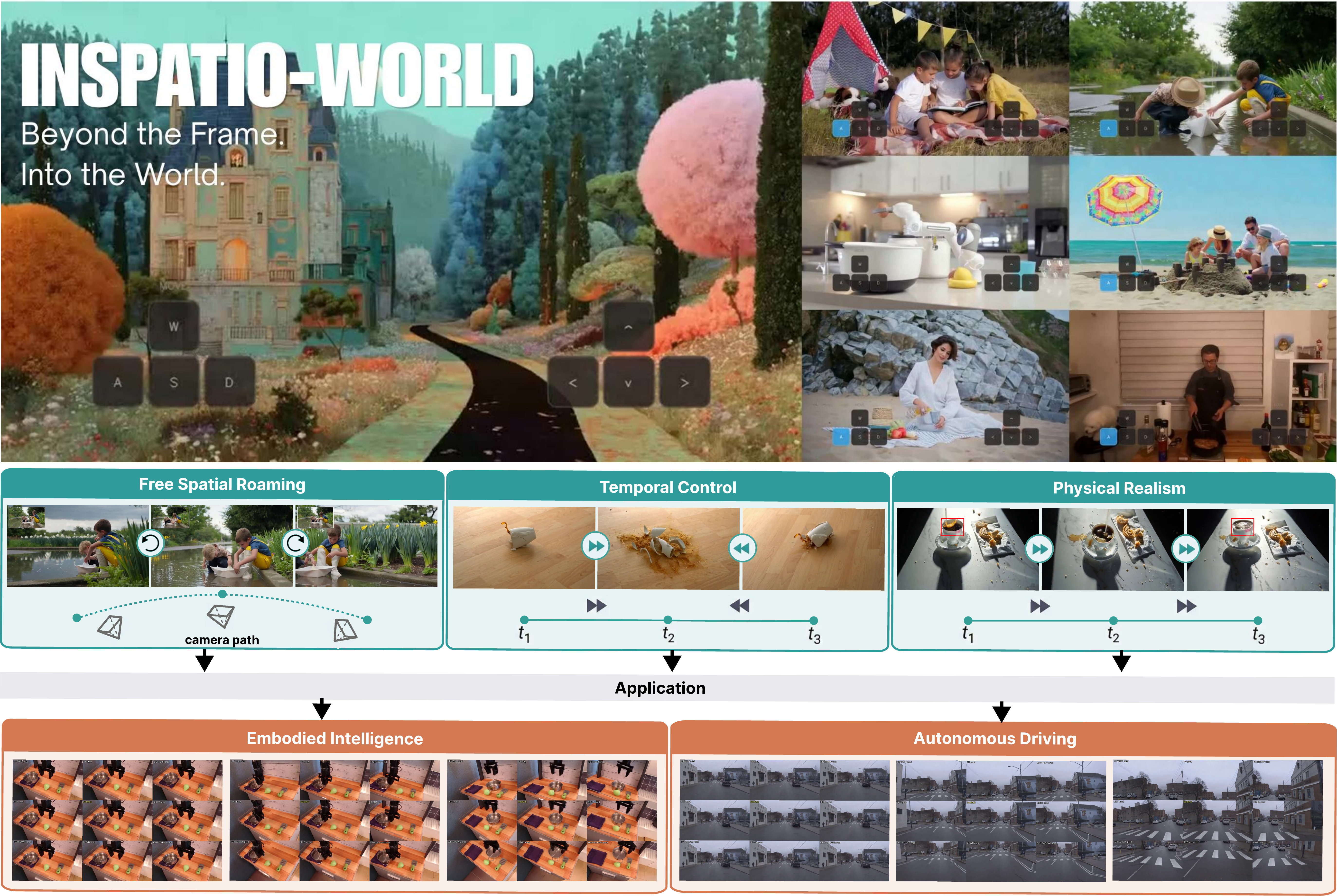}
    \caption{
    \textbf{\method: Toward a Versatile 4D World Simulator.} 
    \textbf{Top:} Our framework enables the synthesis of diverse dynamic scenes from a single video, supporting real-time, high-DoF interactive 4D roaming experiences.
    \textbf{Middle:} The system is driven by those core capabilities: \textit{Free Spatial Roaming} along user-defined camera trajectories, \textit{Temporal Control} over dynamic scene evolution, and the maintenance of \textit{Physical Realism}. 
    \textbf{Bottom:} These capabilities endow \method with the potential to serve as a real-time 4D novel-view rendering engine, promising to support downstream tasks such as \textit{Embodied Intelligence} and \textit{Autonomous Driving}.
    }
    \label{fig:teaser}
\end{figure}

\begin{abstract} 

Building world models with spatial consistency and real-time interactivity remains a fundamental challenge in computer vision. Current video generation paradigms often struggle with a lack of spatial persistence and insufficient visual realism, making it difficult to support seamless navigation in complex environments.
To address these challenges, we propose \method, a novel real-time framework capable of recovering and generating high-fidelity, dynamic interactive scenes from a single reference video. 
At the core of our approach is a Spatiotemporal Autoregressive (STAR) architecture, which enables consistent and controllable scene evolution through two tightly coupled components:
Implicit Spatiotemporal Cache aggregates reference and historical observations into a latent world representation, ensuring global consistency during long-horizon navigation;
Explicit Spatial Constraint Module enforces geometric structure and translates user interactions into precise and physically plausible camera trajectories.
Furthermore, we introduce Joint Distribution Matching Distillation (JDMD). By using real-world data distributions as a regularizing guide, JDMD effectively overcomes the fidelity degradation typically caused by over-reliance on synthetic data.
Extensive experiments demonstrate that \method significantly outperforms existing state-of-the-art (SOTA) models in spatial consistency and interaction precision, ranking first among real-time / interactive methods on the WorldScore-Dynamic benchmark, and establishing a practical pipeline for navigating 4D environments reconstructed from monocular videos.
\end{abstract}

\vspace{10pt}

\begin{minipage}[b]{0.7\textwidth}
\textbf{Project Website:} \href{https://inspatio.github.io/inspatio-world/}{\textcolor{blue}{https://inspatio.github.io/inspatio-world/}}

\textbf{Github:} \href{https://github.com/inspatio/inspatio-world}{\textcolor{blue}{https://github.com/inspatio/inspatio-world}}

\end{minipage}%

\section{Introduction}\label{sec:intro}

Developing world models with spatial consistency and real-time interactivity is a fundamental goal in computer vision.
With recent advances in video diffusion models, the ability to synthesize high-quality dynamic videos from text has demonstrated immense potential for simulating the complexities of the physical world. 
In particular, the rise of interactive video generation has made real-time navigation and dynamic feedback within generated environments possible, laying the foundation for constructing virtual worlds with high degrees of freedom~\citep{genie3,mirage2,sun2025worldplay,bar2025navigation}.

However, despite the ability of existing video diffusion models~\citep{wan2025wan,kong2024hunyuanvideo,videoworldsimulators2024} to synthesize visually striking short clips, they still face fundamental challenges in the task of long-horizon roaming within complex dynamic environments.
Current approaches are primarily limited by the following three bottlenecks:
\begin{enumerate}[leftmargin=*, label=\arabic*.]
\item Spatial Persistence Degradation: Existing autoregressive frameworks lack effective memory mechanisms and explicit geometric guidance, leading to the loss of scene structures and environmental states, or the occurrence of drift, during long-term operation or large viewpoint transitions.
\item Synthetic-to-Real Gap: Due to an over-reliance on synthetic training data, the generated videos exhibit a distribution shift from real-world visual statistics in terms of illumination, textures, and material properties.
\item Insufficient Control Precision: The general inability of current models to accurately execute user-defined trajectories reflects a fundamental deficiency in their underlying spatial geometric reasoning.
\end{enumerate}

To overcome the aforementioned challenges, we propose \method, a novel real-time 4D world model. 
Unlike existing world models ~\citep{genie3,lingbotworld,sun2025worldplay,he2025matrix}, \method is not limited to text and image inputs; instead, it supports transforming a reference video into a "living world" capable of real-time interaction.

The core innovation of this work is two-fold.
At the architectural level, we propose the Spatio-Temporal Autoregressive (STAR) framework. 
This architecture enables the transformation of monocular videos into dynamic, interactive, and immersive navigation experiences, while effectively enhancing spatial consistency and interaction control precision.
Specifically, we develop an implicit spatio-temporal cache that aggregates reference frames and historical generative information within a fixed sliding window. 
This establishes a coupled long-and-short-range memory mechanism, ensuring the temporal stability of long-range generation during real-time exploration.
Building upon this, by introducing explicit spatial constraints, we translate user interactions into precise camera trajectories and seamlessly integrate them into the spatial reasoning process, achieving high-precision camera-controlled generation.
The concept of explicit spatial constraints was initially explored in our prior work, InSpatio-WorldFM~\cite{team2026inspatio}. 
In this study, we generalize it to video generation models and empower it with an optional spatial memory mechanism.

Concurrently, at the learning mechanism level, we propose Joint Distribution Matching Distillation (JDMD) to mitigate the visual appearance degradation inherent in synthetic training data. 
This approach decomposes training into two complementary distillation tasks:
Controllable video rerendering (Video-to-Video, V2V)~\cite{recammaster}, which learns precise motion control and spatiotemporal consistency from synthetic data;
Text-to-Video (T2V) task, which captures text-conditioned generation aligned with real-world data distributions. 
The core mechanism lies in the unified weight-sharing between these two tasks. Gradient guidance extracted from the real-world T2V distribution drives the shared feature space toward alignment and calibration with high-fidelity distributions. 
Consequently, the V2V task maintains high-precision controllability while directly benefiting from the superior texture details and illumination fidelity of the real-world distribution, achieving a synergy between controllable generation and photorealistic quality.
Furthermore, the distinct input structures of the two tasks prevent gradient interference between motion-control learning and visual-fidelity optimization.
As a result, the model optimizes visual quality while strictly adhering to the specified input conditions.

The primary contributions of this work are summarized as follows:
\begin{itemize}[leftmargin=*, itemsep=2pt]
    \item  We introduce \method, 
    a novel real-time framework for spatiotemporal roaming from monocular videos, with publicly released code and models.
    \item We propose a Spatio-Temporal Auto-Regressive (STAR) architecture that leverages an implicit spatio-temporal cache and explicit spatial constraints to achieve high-consistency, high-precision camera control in real-time (Sec. \ref{sec:ar_framework}).
    \item We propose Joint Distribution Matching Distillation (JDMD), a weight-sharing multi-task learning framework that leverages real-world data distributions to guide the feature space alignment of the student model, thereby effectively enhancing the fidelity of the generated regions (Sec. \ref{sec:joint_dmd}).
    \item Extensive quantitative and qualitative evaluations demonstrate that \method significantly outperforms existing generative world models in terms of motion robustness and visual quality. Furthermore, the proposed system achieves real-time performance of 24 FPS while maintaining exceptional spatiotemporal consistency.
\end{itemize}

\section{Related Work}\label{sec:related}

\paragraph{Video diffusion models.}
Video diffusion models have emerged as the prevailing paradigm for high-fidelity video generation~\citep{ho2022video, blattmann2023align, Ho2022ImagenVH, videoworldsimulators2024, polyak2024movie, hacohen2024ltx, blattmann2023stable, villegas2023phenaki, deng2024nova, gupta2024photorealistic}. In recent years, architectures have transitioned from traditional U-Nets~\citep{guo2023animatediff, singer2022make} to more scalable transformer-based designs~\citep{videoworldsimulators2024, hong2023cogvideo, kong2024hunyuanvideo, zheng2024open, wan2025wan}, which unlock superior realism and dynamic fidelity. This foundational progress provides a strong generative backbone for building more complex, interactive spatiotemporal simulations. Among them, Wan2.1~\citep{wan2025wan} demonstrates superior generation capability as an open-source model  and is therefore selected as our backbone.

\paragraph{Novel view synthesis and camera-controllable generation.}
Classical novel view synthesis methods rely on explicit 3D representations such as neural radiance fields~\citep{mildenhall2021nerf} or 3D Gaussian splatting~\citep{kerbl20233d}, which require multi-view input and per-scene optimization.
Recent works have actively explored camera-controllable video generation using diffusion models.
Some approaches~\citep{he2024cameractrl, bahmani2024vd3d, kuang2024collaborative, zheng2024cami2v, liang2025wonderland, xu2024camco, bahmani2025ac3d, wang2024cpa, recammaster} directly inject camera parameters via cross-attention, channel concatenation, or Plücker embeddings.
To provide stronger geometric fidelity and alleviate the cross-modal alignment gap between numerical pose signals and visual content, rendering-based approaches incorporate explicit 3D-aware conditioning by lifting depth to point clouds and using rendered proxy videos. This is seen in methods such as Gen3C~\citep{ren2025gen3c}, MVGenMaster~\citep{cao2025mvgenmaster}, TrajectoryCrafter~\citep{mark2025trajectorycrafter}, and others~\citep{muller2024multidiff, li2025realcam, feng2024i2vcontrol, popov2025camctrl3d, gu2025diffusion, zhai2025stargen, zhang2025recapture, xiao2024trajectory, bian2025gs, yesiltepe2025dynamic}.
Furthermore, several training-free methods have been proposed to achieve flexible camera control~\citep{hou2024training, hu2024motionmaster, ling2024motionclone, xiao2024video}.
For open-ended generation and dynamic scene exploration, methods like Infinite-World~\citep{infiniteworld}, and CameraCtrl II~\citep{he2025cameractrl}, LingBot-World~\citep{lingbotworld}, Google Genie 3~\citep{genie3}, World Labs RTFM~\citep{worldlabs2025rtfm}, Matrix-game 2.0~\cite{he2025matrix} target unbounded horizons.
However, these prior methods fundamentally suffer from spatial persistence degradation due to a lack of effective memory mechanisms and explicit geometric guidance, a synthetic-to-real gap in visual statistics caused by an over-reliance on synthetic training data, and insufficient control precision reflecting a deficiency in underlying spatial geometric reasoning. In contrast, \method systematically overcomes these bottlenecks by injecting reference frames into the KV cache as a global spatiotemporal anchor and utilizing Joint Distribution Matching Distillation to unify explicit 3D constraints with implicit spatial memory and real-world priors, thereby achieving high-fidelity and precisely controllable spatial roaming.

\paragraph{Autoregressive video diffusion.}
Autoregressive formulations have gained traction as a means to enable unbounded-length generation by modeling sequences as step-wise conditionals. Traditional approaches generate spatiotemporal tokens sequentially via next-token prediction~\citep{weissenborn2020scaling, kondratyuk2024videopoet, yan2021videogpt, wang2024loong, Bruce2024GenieGI, ren2025next}.
Recently, hybrid models integrating autoregressive and diffusion frameworks have emerged as a promising direction in the generative modeling of videos and other continuous sequences~\citep{chen2024diffusion, weng2024art, liu2024redefining, guo2025long, hu2024acdit, jin2024pyramidal, gu2025long, gao2024ca2, li2025arlon, liu2024mardini, zhang2025generative, zhang2025test, li2024autoregressive, wu2023ar, deng2024causal, arriola2025block, liu2024autoregressive, zhou2025transfusion, mo2025xfusionintroducingnewmodality}. Additionally, rolling diffusion variants employ progressive noise schedules for sequential generation~\citep{ruhe2024rolling, kim2025fifo, xie2024progressive, zhang2025packing, magi1, sun2025ar}; however, their premature commitment to future frames limits real-time responsiveness to user-injected controls.
Within the autoregressive diffusion paradigm, CausVid~\citep{yin2025causvid} introduces causal attention masks to convert bidirectional models into autoregressive ones, while Self-Forcing~\citep{huang2025selfforcing} bridges the train-test gap to enable streaming generation with KV caching. 
However, they inherently lack the mechanisms to incorporate real-time dynamic control signals, such as continuous camera trajectories or geometric constraints. 
Consequently, they are fundamentally incapable of supporting interactive 4D roaming, as they cannot translate real-time user intentions into deterministic scene exploration. 
To break this limitation, \method explicitly designs a multi-condition autoregressive pathway that seamlessly injects dynamic spatial constraints, transforming passive streaming generation into highly controllable, long-horizon interactive navigation.

\paragraph{Distribution matching distillation.}
The inference efficiency of diffusion models has long been a primary bottleneck limiting their practical application. While Generative Adversarial Networks have recently been repurposed to distill video diffusion models~\citep{zhang2024sf, lin2025diffusion, mao2025osv, wu2025snapgen}, aligning the generated distribution with high-fidelity targets remains a challenge. Early acceleration schemes, such as DDIM or sampler optimizations, yielded promising results but struggled to achieve generation in extremely few steps (e.g., 4-step).
To achieve this, progressive distillation~\cite{salimans2022progressive} gradually compresses the sampling trajectory by halving the number of steps at each stage.
In contrast, consistency models~\cite{luo2023latent} learn a consistency mapping along the ODE trajectory, attempting to reconstruct images from noise in a single step.
The emergence of Distribution Matching Distillation~\citep{yin2024onestep} marks a paradigm shift in distillation. Prior applications, however, have predominantly focused on single-teacher settings. In camera-controlled generation, na\"{\i}vely distilling from a motion-conditioned teacher (typically trained on synthetic data) inevitably forces the student model into a synthetic domain shift, resulting in severe perceptual degradation, texture smoothing, and plastic-like artifacts.
To break this zero-sum game between geometric control and visual quality, we extend DMD to a joint dual-teacher formulation. By synergistically leveraging a perceptual teacher to provide physical prior regularization alongside a motion teacher for precise geometric alignment, \method ensures high-fidelity texture retention without compromising exact camera control.


\begin{figure}[t]
    \centering
    \includegraphics[width=\textwidth]{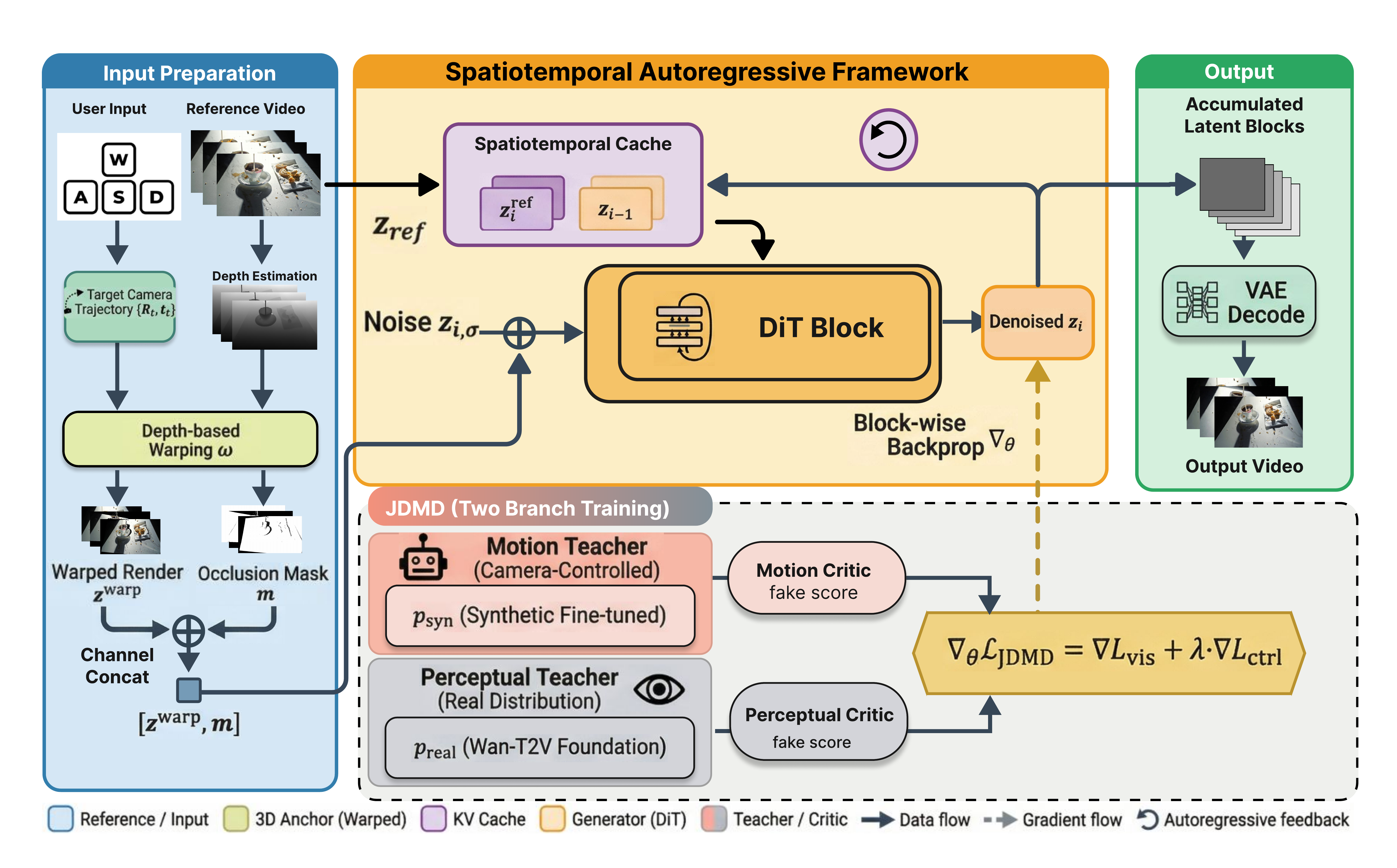}
    \caption{\textbf{Architecture of the Spatiotemporal Autoregressive Framework and JDMD Pipeline.} 
    The framework constructs a spatiotemporal cache using reference information and historical generations, leveraging depth-based warping to establish explicit geometric constraints for consistent autoregressive video generation. 
    The JDMD phase features a multi-task distillation mechanism with shared weights, supervised by a dual-teacher architecture comprising perceptual and motion teachers.
    }
    \label{fig:overview}
\end{figure}

\section{Method}\label{sec:method}

\subsection{Problem Formulation}\label{sec:formulation}

To achieve long-term generation under multimodal constraints, we formulate the generation process as a chunk-wise conditional autoregressive task, where each chunk consists of $K$ consecutive frames. 
Given a global reference context $\mathbf{C}_{\text{ref}}$ and a set of real-time user interaction instructions $\mathcal{T}$, our goal is to model the distribution of the latent sequence $\mathbf{Z}_{1:I}$. 
Following Self-Forcing \cite{huang2025selfforcing}, we apply the probability chain rule to factorize this distribution into a product of stepwise conditional probabilities:
\begin{equation}\label{eq:formulation}
    p(\mathbf{Z}_{1:I} \mid \mathbf{C}_{\text{ref}}, \mathcal{T}) = \prod_{i=1}^{I} p(\mathbf{z}_i \mid \mathbf{z}_{<i}, \mathbf{c}^{\text{ref}}_i, \tau_i),
\end{equation}
where the generation of the $i$-th block $\mathbf{z}_i$ is jointly constrained by the historical context $\mathbf{z}_{<i}$, the reference guidance $\mathbf{c}^{\text{ref}}_i$, and the interaction term $\tau_i$.

\subsection{Spatiotemporal Autoregressive Framework}\label{sec:ar_framework}

To ensure spatial persistence and interactive precision during long-horizon interactive roaming, we propose a spatio-temporal autoregressive framework, as illustrated in Fig.~\ref{fig:overview}. The framework comprises two key components:
First, by aggregating historical and reference frames to construct an implicit ST-Cache, the framework leverages short-term historical memory and long-term reference information to jointly guide the generation process, thereby maintaining temporal continuity and spatial consistency.
Second, by incorporating the geometric information of reference frames to enhance multi-view consistency, the system transforms user control commands into explicit spatial constraints, achieving precise camera control.
Ultimately, the system synergistically injects the implicit memory states and explicit geometric constraints into the Diffusion Transformer (DiT), enabling high-fidelity, real-time generation of interactive dynamic environments.

Under this framework, the denoising process for generating the $i$-th block $\mathbf{z}_i$ can be expressed as:
\begin{equation}\label{eq:denoise}
    \hat{\mathbf{z}}_i = \text{Denoise}_\theta(\mathbf{z}_{i, \sigma} \mid \mathbf{z}_{<i}, \mathbf{z}^{\text{ref}}_i, [\mathbf{z}^{\text{warp}}_i, \mathbf{m}_i]),
\end{equation}
where $\mathbf{z}_{i, \sigma}$ is the initial latent of the $i$-th block at noise level $\sigma$. The model is synergistically constrained by three types of conditions:
\begin{itemize}[leftmargin=*,itemsep=2pt]
    \item Historical condition ($\mathbf{z}_{<i}$): The generated latent of previous blocks. It carries the local temporal context, ensuring motion smoothness and logical continuity between blocks.
    \item Reference condition ($\mathbf{z}^{\text{ref}}_i$): The corresponding latents retrieved and compressed from the reference video in real time. Serving as a global spatial anchor, it ensures that the model can accurately trace back the textures and semantic features of the original scene even after long-horizon roaming.
    \item Geometric condition ($[\mathbf{z}^{\text{warp}}_i, \mathbf{m}_i]$): The explicit constraint driven by the current interaction instruction $\tau_i$. Here, $\mathbf{z}^{\text{warp}}_i$ represents the geometrically aligned reprojection features, and $\mathbf{m}_i$ is the valid pixel mask. Together, they provide deterministic spatial structural guidance to prevent scene distortion.
\end{itemize}

\subsubsection{Spatiotemporal Cache Mechanism with Differentiable Recomputation}\label{sec:cache_mechanism}

To effectively mitigate the state drift that is common in autoregressive generation and to meet the demands of interactive real-time inference, we propose a spatiotemporal cache mechanism. 
The essence of this mechanism is to integrate short-term temporal information (historical frames) with long-term spatiotemporal anchors (reference frames), achieving high-fidelity end-to-end content generation with constant KV cache memory overhead.
Specifically, when generating the $i$-th block, the system retrieves the corresponding latent $\mathbf{z}^{\text{ref}}_i$ from the reference video to serve as a globally stable spatiotemporal anchor.
Meanwhile, to ensure the smoothness of motion, the previously generated latent $\mathbf{z}_{i-1}$ is organized as a sliding window and stored in the cache, which prevents memory overflow during long-sequence inference while maintaining local temporal continuity. 

Furthermore, to address the distribution shift caused by the growth of the sequence length in Rotary Position Embedding (RoPE) during long-horizon inference, we adopt a position index fixing strategy. By anchoring the starting position indices of the current block $\mathbf{z}_i$, the reference anchor $\mathbf{z}^{\text{ref}}_i$, and the historical block $\mathbf{z}_{i-1}$ to a preset absolute coordinate origin (denoted as $f_i$, $f^r_i$, and $f^h_i$, respectively), we constrain the receptive field of the model within a stable representation space. This relative pose-fixed encoding eliminates the numerical instability arising from temporal extrapolation and assists the noisy latent in building stable correlations with the reference and the historical contexts, thereby significantly enhancing spatial consistency.

In addition, to address the differentiability requirements and memory bottlenecks during training, we propose a Chunk-wise Backpropagation strategy. Existing autoregressive diffusion models often resort to gradient-free modes for KV Cache construction when computing distribution losses (e.g., DMD Loss), due to the prohibitive memory pressure as the sequence length increases. 
Such non-differentiability forces the model into passive feature fitting, thereby constraining the overall generation quality.
The proposed strategy decouples forward inference from backward optimization, reducing peak memory usage to the scale of a single chunk. The procedure consists of two stages:
In Stage 1, a full-length inference is performed in gradient-free mode, retaining only the final output to compute the DMD loss. This captures global supervisory signals with negligible computational overhead.
In Stage 2, the forward pass is re-executed chunk-by-chunk to trigger backpropagation. This process encompasses the entire pipeline—including KV Cache construction and denoising—while intermediate representations are released immediately following each gradient update.
This time-space tradeoff strategy ensures full-link differentiability within each chunk, enabling the model to precisely learn more expressive spatiotemporal features and significantly enhancing generation fidelity.

\subsubsection{Geometry-Aware Explicit Constraints}\label{sec:geometry_constraints}

To respond precisely to dynamic interaction instructions $\tau_i$, we introduce an explicit geometric constraint mechanism that translates discrete user operations into deterministic spatial structural guidance. This process consists of two stages: pose evolution and geometric feature projection. First, the system maps the user's rotation, translation, and perspective shift instructions for the current block into a 6-Degree-of-Freedom (6-DoF) relative pose transformation $\Delta \mathbf{T}_i$. The global pose $\mathbf{T}_i$ corresponding to the $i$-th block is defined as the accumulation of all historical interactions, derived recursively by applying $\Delta \mathbf{T}_i$ to the previous camera state $\mathbf{T}_{i-1}$.

After obtaining the current pose $\mathbf{T}_i$, the system geometrically aligns the reference features with the current viewpoint using a projection function. Specifically, the Feed-Forward Reconstruction (FFR) methods~\citep{vggt2025,pi3geometry2026,depthanythingv3} are employed to extract geometric priors from the reference video latents, yielding a depth map $\mathbf{D}_{\text{ref}}$ and camera intrinsics $\mathbf{K}$. Based on $\mathbf{T}_i$, the system executes the following reprojection operation:
\begin{equation}\label{eq:projection}
    \mathbf{z}^{\text{warp}}_i, \mathbf{m}_i = \text{Proj}(\mathbf{z}^{\text{ref}} \mid \text{FFR}(\mathbf{z}^{\text{ref}}), \mathbf{T}_i),
\end{equation}
where $\mathbf{z}^{\text{warp}}_i$ represents the geometrically aligned guidance feature. 
To effectively distinguish between black texture and invisible regions, we concatenate a binary mask $\mathbf{m}_i$ to the latent representation. 
By explicitly defining the valid reprojection regions, this mask guides the autoregressive model to generate under deterministic structural constraints. 

Furthermore, by natively supporting the injection of geometric constraints, our model enables an optional explicit structural memory mechanism. 
By reconstructing the generated video and dynamically expanding the point-cloud map, the system constructs a structured representation of the scene with minimal computational overhead. This explicit geometric constraint effectively functions as a spatial memory proxy, providing a fundamental structural anchor for long-range generation.

\subsubsection{Multi-Condition Causal Initialization}\label{sec:causal_init}
 
In the field of autoregressive video generation, a well-designed initialization strategy is a critical prerequisite for ensuring training convergence stability and sequence consistency. 
Prevailing frameworks, represented by CausVid~\cite{yin2025causvid}, typically initialize the student model with causal attention masking to enforce a causal generative paradigm in which the synthesis of the current frames is strictly conditioned on the preceding generative context.

However, this initialization strategy, which relies on causal attention mask, exhibits notable deficiencies in multi-condition controllable generation. Since the synthesis of each chunk must integrate heterogeneous inputs—including preceding frames, reference images, and geometric constraints—simple causal masks are inadequate for modeling the intricate causal interplays among these disparate signals. Consequently, directly applying this paradigm often leads to suboptimal generative quality.

To address these challenges, we proposes a Multi-conditional Causal Initialization strategy. 
Deviating from traditional static causal masking, 
this strategy performs chunk-wise autoregressive multi-step rehearsal directly on ground-truth data or teacher-model ODE trajectories, 
ensuring the model establishes accurate associations with various conditions during the initial phase. 
In the subsequent distillation phase, with robust causal dependencies already established, the student model shifts its focus to sampling acceleration (multi-to-few steps) and fidelity refinement (coarse-to-fine details).

Furthermore, explicit geometric constraints injected via channel concatenation are confined to the current denoising block. By applying zero-padding to the corresponding channels of historical blocks, we ensure the history cache provides only pure image information. 
This design prevents the infiltration of past geometric signals, safeguarding the integrity of the controlled spatiotemporal autoregressive process and the robustness of the generative logic.

\subsection{Joint Distribution Matching Distillation}\label{sec:joint_dmd}

The realization of interactive roaming tasks depends heavily on the precise decoupling of visual continuity and motion feedback. 
However, the training process supporting reference video inputs requires multi-view synchronized video streams, and such high-fidelity annotated data is extremely scarce in real-world scenarios. 
Although synthetic data provide perfect geometric constraints, the inherent domain shift of synthetic data often leads to perceptual degradation phenomena, such as texture smoothing and structural repetition. 
To circumvent the intrinsic trade-off between controllability and visual fidelity, we propose Joint Distribution Matching Distillation (JDMD).

We first briefly recap the fundamental principles of Distribution Matching Distillation (DMD)~\cite{yin2024onestep}. 
Standard DMD trains a student generator to match the distribution of a teacher diffusion model by minimizing the Kullback-Leibler (KL) divergence. 
The gradient of the student model's parameters is given by:
\begin{equation}\label{eq:dmd_grad}
    \nabla_\theta \E_t\!\left[D_{\mathrm{KL}}\!\left(p_{\theta,t} \| p_{\mathrm{data},t}\right)\right]
    = -\E_{t,\,\hat{\x}_t}\!\left[\left(s_{\mathrm{real}}(\hat{\x}_t, t) - s_{\mathrm{fake}}(\hat{\x}_t, t)\right) \frac{\partial \hat{\x}}{\partial \theta}\right],
\end{equation}
where $s_{\mathrm{real}}$ and $s_{\mathrm{fake}}$ are the score functions approximated by the real (teacher) and the fake (student-tracking) score networks, respectively, and $\hat{\x}_t$ is the noisy version of the output of the student model.

The core idea of JDMD is to employ a multi-task learning paradigm that leverages real-world data distributions as a regularization guidance to overcome the fidelity degradation inherent in synthetic data.
Specifically, JDMD synergistically guides the student model using two frozen teacher distributions by alternately activating two distillation tasks during training iterations: 
in the controllable video rerendering (V2V) task, the student model receives the reference video and geometric information to focus on learning precise motion control and spatio-temporal consistency, where the synthetic data distribution $p_{syn}$ is represented by a teacher model fine-tuned on synthetic data to compute the conditional control loss $\mathcal{L}_{\text{ctrl}}$; 
meanwhile, in the Text-to-Video (T2V) task, the student model operates solely conditioned on text to focus on capturing the fidelity and richness of real-world data, where the real-world data distribution $p_{real}$ is represented by the original Wan-T2V foundation model to compute the vision distillation loss $\mathcal{L}_{\text{vis}}$. 
By combining these two objectives, the overall loss function is formulated as a weighted sum:
\begin{equation}\label{eq:joint_dmd_total}
    \Loss_{\text{JDMD}} = \Loss_{\text{vis}} + \lambda_{\text{ctrl}} \Loss_{\text{ctrl}},
\end{equation}
where $\lambda_{\text{ctrl}}$ is a hyperparameter that balances the weights of visual fidelity and motion control.

This dual-track distillation mechanism ensures that when the student model receives an interaction command $\tau$ and a reference video, the condition adherence learned from the controllable V2V task plays a dominant role, guaranteeing precise camera movement and spatio-temporal consistency in the generated output.
Concurrently, the distillation process of the T2V task performs a critical distribution calibration by aligning the feature space with the real-world data distribution, significantly enhancing the visual fidelity of the generated output.
Through Joint Distribution Matching Distillation, \method successfully balances motion compliance with visual fidelity: while maintaining native high-fidelity image quality, the model achieves precise adherence to both reference videos and complex camera trajectories. This mechanism enables the system to ultimately break through the distribution limits of synthetic data, achieving an effective balance between spatial consistency and visual realism in interactive roaming tasks.


\subsection{Implementation Details}

Our training framework leverages diverse data sources, encompassing large-scale publicly available internet videos such as RealEstate10K\cite{re10k}, as well as synthetic datasets specifically tailored for novel-view video rerendering tasks. 
The latter includes both Unreal Engine (UE) rendered sequences and the publicly accessible ReCamMaster\cite{recammaster} dataset.
For each video clip, we apply a feedforward reconstruction model to estimate depth information.
The training procedure follows the Self-Forcing paradigm~\citep{huang2025selfforcing}, with  Wan2.1~\cite{wan2025wan} as the backbone.
The training procedure is divided into three stages, focusing on learning rate scheduling rather than iteration counts:
\begin{itemize}
    \item \textbf{Teacher Training:} The teacher model is trained to establish a robust performance baseline with a learning rate of $2 \times 10^{-5}$.
    \item \textbf{Initialization Phase:} The student model undergoes an initialization stage to establish its auto-regressive inference capability, employing a learning rate consistent with that of the teacher training phase.
    \item \textbf{Student Distillation (JDMD):} The student model is trained under the supervision of the pre-trained teacher. 
    In this stage, the learning rates for the student network and the fake score discriminator are set to $4.0 \times 10^{-6}$ and $8.0 \times 10^{-7}$, respectively.
\end{itemize}

To improve inference efficiency, we employ two acceleration strategies. First, we replace the original Wan-VAE with a lightweight Tiny-VAE ~\cite{BoerBohan2025TAEHV}. 
Although this substitution introduces a slight performance degradation, it offers a favorable trade-off for low-latency real-time applications. 
Second, while the distilled model already achieves efficient inference, we further reduce runtime overhead using graph-level compilation optimizations (using \textit{torch.compile}), which brings additional practical speedup. 
Combined with a model architecture that is naturally compatible with streaming inference, these optimizations enable \method (1.3B model) to achieve a real-time inference speed of 24 FPS on an H-series NVIDIA GPU, and maintain a highly competitive 10 FPS on a consumer-grade RTX 4090 GPU. This demonstrates the framework's broad suitability for interactive applications across varying hardware constraints.

\section{Experiments}\label{sec:experiments}

\subsection{Experimental Setup}\label{sec:setup}

We evaluate the effectiveness of \method through three complementary tasks: 
\begin{itemize}[leftmargin=*,itemsep=2pt]
\item \textbf{WorldScore Benchmark} ~\cite{duan2025worldscore}, evaluates a model's performance in next-scene generation by measuring the precision of instruction control, the stability of spatial structures, and the authenticity of physical dynamics; 
\item  \textbf{Long-term Image-to-Video Generation}, which employs RealEstate10K (RE10K)~\cite{re10k} to examine the model's performance in long-range camera control, content distribution consistency, and visual quality through the generation of long-sequence videos; 
\item  \textbf{Camera Controlled Generative Video Rerendering}, evaluated on both real-world~\cite{nan2024openvid} and synthetic datasets (from PostCam~\cite{chen2025postcam}) to test camera control precision, generation quality, and adherence to original video conditions under given reference video constraints. 
\end{itemize}

In the WorldScore evaluation, we strictly adhere to the official recommendations by adopting the full set of its 10 defined core evaluation metrics.
For the long-term I2V and video rerendering tasks, we have constructed a multi-dimensional and comprehensive quantitative evaluation framework:
\begin{itemize}[leftmargin=*,itemsep=2pt]
\item \textbf{Control Accuracy}, which quantifies the precision of camera motion control by calculating rotation error ($Rot$) and translation error ($Trans$) between the generated sequences and preset trajectories; 
\item \textbf{Generative Distribution Quality}, which uses FID and FVD to measure the similarity between the generated results and real data distributions from image and video perspectives, respectively; 
\item \textbf{Visual Quality}, which encompass six key dimensions of VBench~\cite{huang2024vbench}: Aesthetic Quality, Image Quality, Temporal Flickering, Motion Smoothness, Subject Consistency, and Background Consistency.
\end{itemize}
To comprehensively validate performance, we compare \method against state-of-the-art methods across different technical trajectories, including WorldScore evaluation models such as FantasyWorld~\cite{fantasyworld}, TeleWorld~\cite{chen2025teleworld}, and industrial-grade models like CogVideoX-I2V~\cite{hong2023cogvideo}, Gen-3~\cite{gen3}, LTX-Video~\cite{ltxvideo}, and Hailuo~\cite{hailuo}; 
open-source world models including Infinite-World~\cite{infiniteworld}, LingBot-World~\cite{lingbotworld}, and HY-WorldPlay~\cite{sun2025worldplay}; 
and generative video rerendering baselines such as TrajectoryCrafter~\cite{trajectorycrafter}, ReCamMaster~\cite{recammaster}, and NeoVerse~\cite{neoverse}.

\input{table/worldscore}
 
\subsection{WorldScore Benchmark}\label{sec:worldscore}

We conduct a comprehensive evaluation of \method on the WorldScore benchmark. As shown in Table~\ref{tab:worldscore} and Fig.~\ref{fig:worldscore}, \method (1.3B) achieves state-of-the-art (SOTA) performance in both metrics and computational efficiency among all real-time/interactive methods. Quantitative analysis (Table~\ref{tab:worldscore}) demonstrates that \method outperforms existing real-time baselines in key metrics, notably in overall dynamic score (68.72) and camera control accuracy (81.51). The strong overall dynamic performance and precise control validate the superiority of the spatiotemporal autoregressive framework, while the high generation quality confirms the improvement brought by JDMD. Notably, while maintaining these excellent results, our generation speed is at the forefront; to the best of our knowledge, it is the only world model on the leaderboard capable of reaching 24 FPS real-time operation.

\input{table/i2v}
\subsection{Long-term Image-to-Video Generation}\label{sec:i2v}

Long-horizon generation is a critical task for evaluating interactive world models, as it requires the model to maintain \textit{spatial persistence} and suppress \textit{kinetic drift} and error accumulation over extended sequences. 
We established a rigorous evaluation benchmark by randomly selecting 100 sequences exceeding 150 frames from the RE10K dataset~\cite{re10k}.
Under identical input conditions, we compared \method with state-of-the-art (SOTA) world models. 
For a fair comparison, we employ the 14B version to maintain consistency with LingBot-World.

As shown in Table~\ref{tab:i2v}, \method achieves substantial improvements across all metrics. In terms of generation quality, it yields an FID of 42.68 and an FVD of 100.55, substantially outperforming existing SOTA methods. Most notably, regarding camera motion accuracy, \method demonstrates an overwhelming advantage, with its trajectory error being significantly lower than that of the runner-up, LingBot-World~\cite{lingbotworld}. This numerical dominance establishes our framework's superiority in handling complex, long-duration interactive roaming tasks.

Qualitative results (see Fig.~\ref{fig:i2v_qualitative}) further illuminate the distinct failure modes of baseline methods during extended generation:
Infinite-World~\cite{infiniteworld} suffers from severe structural distortion and geometric warping as the sequence length increases;
HY-WorldPlay~\cite{sun2025worldplay} exhibits a lack of robust motion control, often degenerating into static frame generation;
LingBot-World~\cite{lingbotworld}, while preserving per-frame visual quality, fails to precisely follow intended trajectories due to inaccurate camera pose estimation.
In contrast, by incorporating a global spatial reference, \method ensures the geometric integrity of the scene and maintains precise camera control, enabling artifact-free long-horizon navigation.

\input{table/v2v}
\subsection{Camera Controlled Generative Video Rerendering}\label{sec:v2v}

To evaluate the performance of \method on the task of generative video rerendering under camera control, we conducted experiments on both the synthetic Blender dataset and the real-world OpenVid dataset. 
The Blender evaluation set consists of 100 samples, each featuring precise trajectories and ground-truth videos. 
The OpenVid evaluation set contains 240 samples, constructed by matching 40 original OpenVid videos with 6 complex trajectories in different directions.
Since the videos of OpenVid lack corresponding ground-truth target videos for calculating distribution discrepancies, we employ VBench to evaluate the video generation quality.
For a fair comparison, we employ the 14B version to maintain consistency with Neoverse.

Quantitative results demonstrate that our approach achieves state-of-the-art (SOTA) performance on both datasets (see Table~\ref{tab:v2v_combined}). 
Specifically, \method  outperforms existing methods in FID, FVD, and comprehensive video quality metrics, while achieving comparable camera control accuracy to current SOTA models. This firmly demonstrates the effectiveness of the proposed method.
Furthermore, qualitative evaluations (see Fig.~\ref{fig:v2v_qualitative}) visually highlight the advantages of our approach. Compared to other methods, \method exhibits superior video generation quality. 
Notably, although Neoverse demonstrates good generation quality and camera control accuracy, it exhibits limited capacity in preserving spatio-temporal coherence relative to the input video, resulting in inferior FID and FVD scores. 
In contrast, our method strictly preserves high consistency with the input reference video while achieving high-quality generation.
Finally, to the best of our knowledge, \method is currently the only open-source generative video rerendering solution capable of real-time execution.

\section{Discussion and Conclusions}\label{sec:conclusion}

In this technical report, we introduce \method, an innovative 4D generative world model specifically engineered for real-time interactive roaming.
By constructing an efficient spatio-temporal autoregressive framework, we successfully integrate an implicit ST-Cache for long-term spatio-temporal anchoring with explicit spatial constraints. 
The proposed framework effectively mitigates the critical challenges of spatial persistence loss and imprecise control inherent in interactive video generation. 
To further enhance visual quality, we propose Joint Distribution Matching Distillation (JDMD), which utilizes a dual-teacher paradigm to decouple and simultaneously optimize motion fidelity and perceptual realism, effectively bridging the domain gap between synthetic simulation and physical reality. Experimental results demonstrate that the proposed framework establishes a new state-of-the-art in spatial continuity and visual precision while maintaining high-efficiency performance at 24 FPS, providing a robust foundation for high-degree-of-freedom navigation in synthesized virtual worlds.

\subsection{Limitation}
Despite the significant advancements of \method, the system exhibits certain limitations in maintaining long-term consistent memory of generated regions and enabling seamless 360-degree dynamic roaming.
Specifically, while our framework successfully integrates external spatio-temporal anchors and explicit point-cloud memory to uphold spatial consistency, it primarily functions as a structural backbone that falls short of persistently encoding the fine-grained textural details of autonomously generated areas.
Furthermore, while this explicit geometric scheme effectively supports large-scale displacement in static environments, ensuring the multi-view consistency and spatio-temporal coherence of dynamic elements during wide-angle, omnidirectional view transitions remains an open challenge.

\subsection{Future Work}
Looking ahead, we will focus on developing a more profound semantic memory system, exploring the deep coupling of geometric structures with high-dimensional textural features to achieve comprehensive, full-spatio-temporal recording and reconstruction of generated regions.
Concurrently, we intend to investigate long-range dynamic constraint mechanisms by introducing stronger physical priors into the autoregressive process. 
Our goal is to achieve perfect closed-loop simulation of large-scale, high-complexity dynamic scenes under physical guidance, continuously pushing generative world models toward higher dimensions and broader application horizons.


\section*{Acknowledgment}
The authors are deeply grateful to thank Chaoran Tian, Gan Huang, Hengxu Lin, Jingbo Liu, and Zhiwei Huang for their valuable support and assistance throughout this research.

{\small
\bibliographystyle{plainnat}
\bibliography{refs}
}

\end{document}

%% file: table/worldscore.tex
\begin{table*}[!t]
\centering
\caption{\textbf{WorldScore benchmark results.} We compare \method against leading world models on the WorldScore benchmark. Our method achieves the highest camera control score and highly competitive overall dynamic performance at a fraction of the computational cost. The best results are highlighted in bold, and the second-best are \underline{underlined}.}
\label{tab:worldscore}
\resizebox{\textwidth}{!}{%
\begin{tabular}{l c cccc cc ccc}
\toprule
\multirow{2}{*}{Method} & \multirow{2}{*}{Real-time/Interactive} & \multicolumn{4}{c}{Dynamic Score} & \multicolumn{2}{c}{Control} & \multicolumn{3}{c}{Static Score} \\
\cmidrule(lr){3-6} \cmidrule(lr){7-8} \cmidrule(lr){9-11}
 &  & Overall$\uparrow$ & 3D Consist.$\uparrow$ & Motion Acc.$\uparrow$ & Smoothness$\uparrow$ & Camera$\uparrow$ & Object$\uparrow$ & Overall$\uparrow$ & Photometric$\uparrow$ & Content$\uparrow$ \\
\midrule
FantasyWorld-1.0  & No  & \textbf{71.39} & 84.62          & 50.30          & \textbf{75.81} & \underline{81.45} & \textbf{87.90} & \textbf{80.45} & \textbf{94.07} & 66.94          \\
CogVideoX-I2V     & No  & 59.12          & \underline{86.21} & \underline{69.56} & 60.15          & 38.27          & 40.07          & 62.15          & 88.12          & 36.73          \\
Gen-3             & No  & 57.58          & 68.31          & 54.53          & 68.87          & 29.47          & 62.92          & 60.71          & 87.09          & 50.49          \\
LTX-Video         & No  & 56.54          & 78.41          & \textbf{76.22} & 71.09          & 25.06          & 53.41          & 55.44          & 88.92          & 39.73          \\
Hailuo            & No  & 56.36          & 67.18          & 63.46          & 70.07          & 22.39          & 69.56          & 57.55          & 62.82          & \textbf{73.53} \\
\midrule
TeleWorld         & Yes & 66.73          & \textbf{87.35} & 53.94          & 34.18          & 76.58          & \underline{74.44} & \underline{78.23} & 88.82          & \underline{73.20} \\
\method           & Yes & \underline{68.72} & 84.18          & 60.21          & \underline{71.91} & \textbf{81.51} & 71.63          & 75.81          & \underline{93.00} & 54.50          \\
\bottomrule
\end{tabular}%
}
\end{table*}

\begin{figure}[!t]
    \centering
    \includegraphics[width=0.6\textwidth]{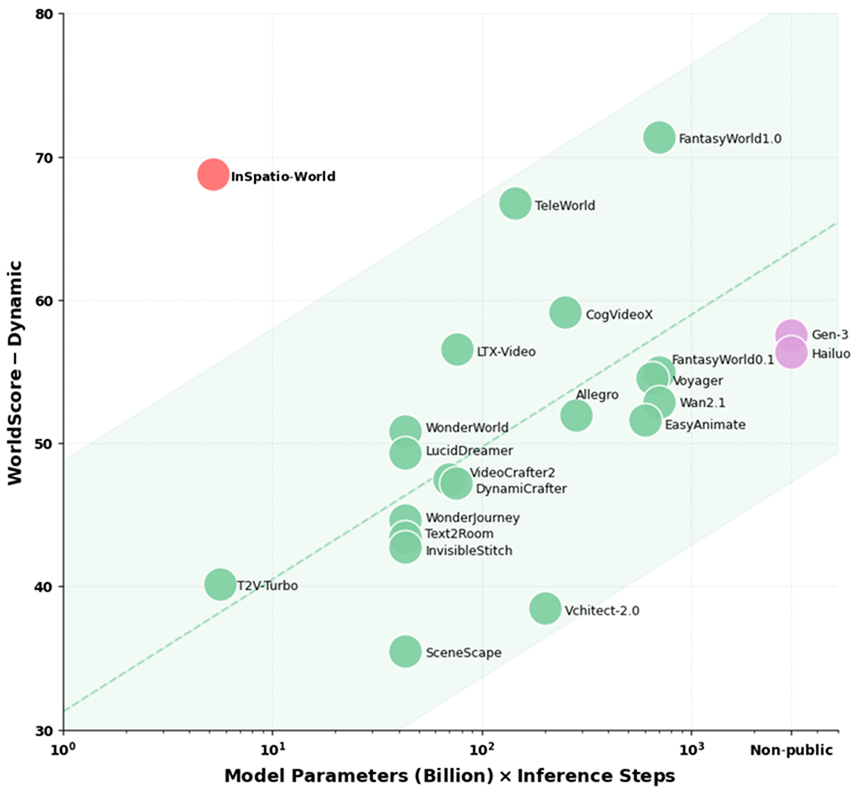}
    \caption{\textbf{Quantitative comparison on WorldScore-Dynamic.} Each bubble represents a method, with the vertical axis showing the score of WorldScore-Dynamic and the horizontal axis showing model parameters $\times$ inference steps. \method achieves a dynamic score of 68.72 with a significantly lower computational overhead, demonstrating a superior compute-quality trade-off by breaking the zero-sum game between geometric control and generation fidelity.}
    \label{fig:worldscore}
\end{figure}

%% file: table/i2v.tex
\begin{table}[!t]
    \centering
    \caption{\textbf{Quantitative comparison on the RE10K-Long dataset.} 
    The best results are highlighted in bold, and the second-best are \underline{underlined}.}
    \label{tab:re10k_long}
    \resizebox{0.6\columnwidth}{!}{%
    \begin{tabular}{l cc cc}
    \toprule
    Method & FID$\downarrow$ & FVD$\downarrow$ & Rot$\downarrow$ & Trans$\downarrow$ \\
    \midrule
    HY-WorldPlay    & 129.46         & 387.50          & 25.050          & 0.6725 \\
    Infinite-World  & 89.44          & 215.96          & 16.518          & 0.4715 \\
    LingBot-World   & \underline{64.84}  & \underline{173.02} & \underline{11.981} & \underline{0.2064} \\
    \midrule
    \method          & \textbf{42.68} & \textbf{100.55} & \textbf{2.8762} & \textbf{0.1398} \\
    \bottomrule
    \end{tabular}%
    }
    \label{tab:i2v}
\end{table}

\begin{figure}[!t]
    \centering
    \includegraphics[width=\textwidth]{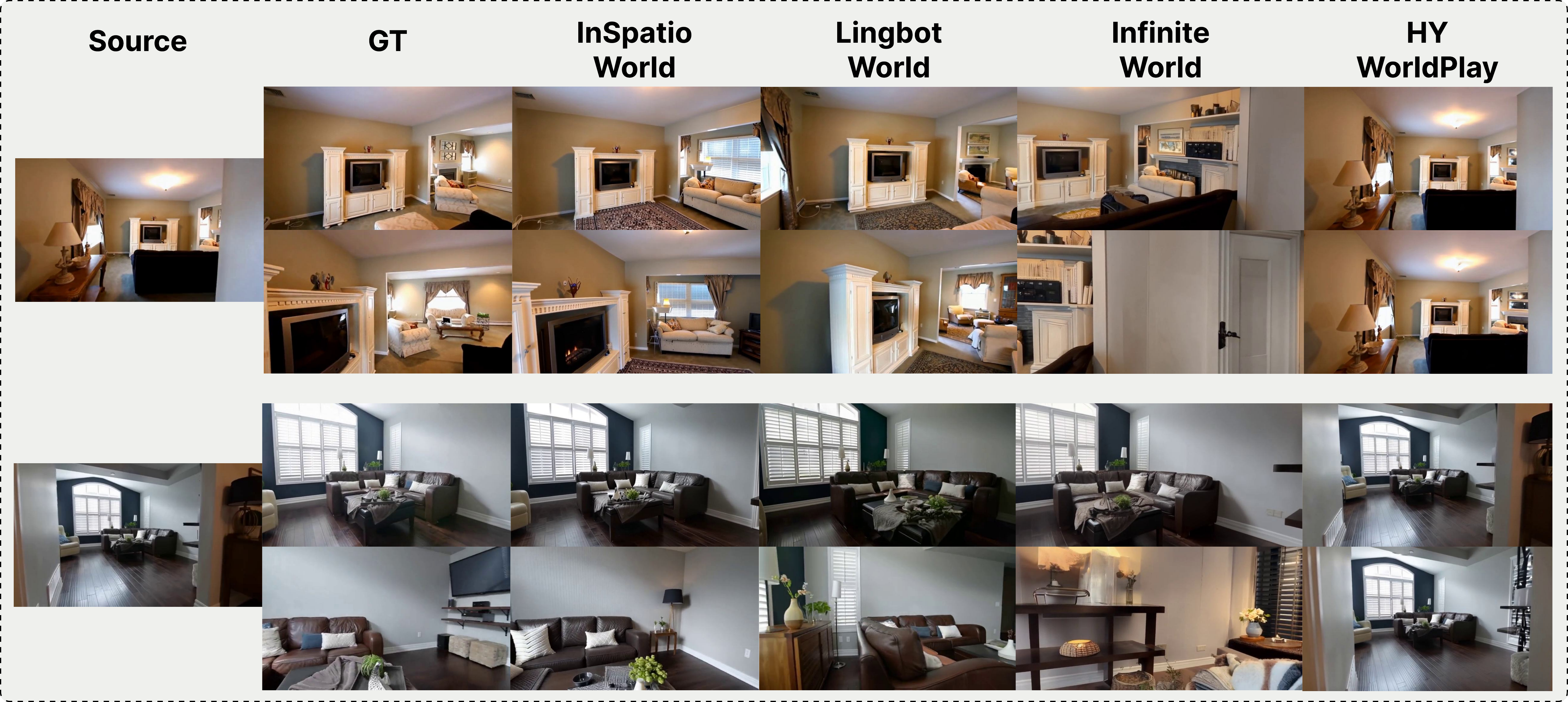}
    \caption{\textbf{Qualitative comparison on RE10K-Long dataset.} 
    Qualitative comparison on RE10K-Long. 
    For each of the two scenes, the leftmost image represents the input Source image. 
    For each method, the top row displays the intermediate frame of the generated sequence, while the bottom row showcases the final frame. 
    As generation progresses, baseline methods exhibit varying degrees of failure, such as camera pose drift or structural warping. 
    In contrast, \method maintains precise trajectory control and persistent geometric consistency throughout the extended sequence.
    }
    \label{fig:i2v_qualitative}
\end{figure}

%% file: table/v2v.tex
\begin{table*}[!t]
\centering
\caption{\textbf{Quantitative comparison on Camera Controlled Video Rerendering.} We evaluate our method against state-of-the-art baselines on both the OpenVid dataset and the synthetic Blender dataset. The best results are highlighted in bold, and the second-best are \underline{underlined}.
For OpenVid dataset, \textbf{Overall} represents the average score of the six VBench metrics.}
\label{tab:v2v_combined}
\resizebox{\textwidth}{!}{%
\begin{tabular}{l | cccccc c | cc | cccc}
\toprule
\multirow{3}{*}{Method} & \multicolumn{9}{c}{OpenVid} & \multicolumn{4}{c}{Blender} \\
\cmidrule(lr){2-10} 
\cmidrule(lr){11-14}
 & \multicolumn{7}{c}{VBench $\uparrow$} &
 \multirow{2}{*}{Rot $\downarrow$}&
 \multirow{2}{*}{Trans $\downarrow$}
 & \multirow{2}{*}{FID $\downarrow$} & \multirow{2}{*}{FVD $\downarrow$} & \multirow{2}{*}{Rot $\downarrow$} & \multirow{2}{*}{Trans $\downarrow$} \\
\cmidrule(lr){2-8} 
 & Aesth. & Imag. & Flick. & Smooth. & Subj. & Bg. & \textbf{Overall} & & & & & & \\
\midrule
TrajectoryCrafter & 0.5210 & 0.6527 & 0.9444 & 0.9736 & 0.8749 & 0.8961 & 0.8105 & 2.1650 & 0.1710 & 256.69 & 818.73 & 4.1780 & 0.2015 \\
ReCamMaster       & 0.5666 & 0.6863 & \textbf{0.9736} & \textbf{0.9928} & \textbf{0.9373} & 0.9163 & 0.8455 & 3.8640 & 0.2310 & 116.53 & 311.06 & 3.5062 & 0.2001 \\
NeoVerse          & \underline{0.5583} & \underline{0.7272} & \underline{0.9646} & \underline{0.9904} & \underline{0.9234} & \textbf{0.9279} & \underline{0.8486} & \textbf{1.5780} & \underline{0.1340} & \underline{103.23} & \underline{230.87} & \textbf{1.2148} & \textbf{0.0636} \\
\midrule
\method           & \textbf{0.5742} & \textbf{0.7296} & 0.9638 & 0.9901 & 0.9216 & \underline{0.9249} & \textbf{0.8507} & \underline{1.6000} & \textbf{0.1240} & \textbf{44.46} & \textbf{110.11} & \underline{1.2386} & \underline{0.0667} \\
\bottomrule
\end{tabular}%
}
\end{table*}

\begin{figure}[!t]
    \centering
    \includegraphics[width=\textwidth]{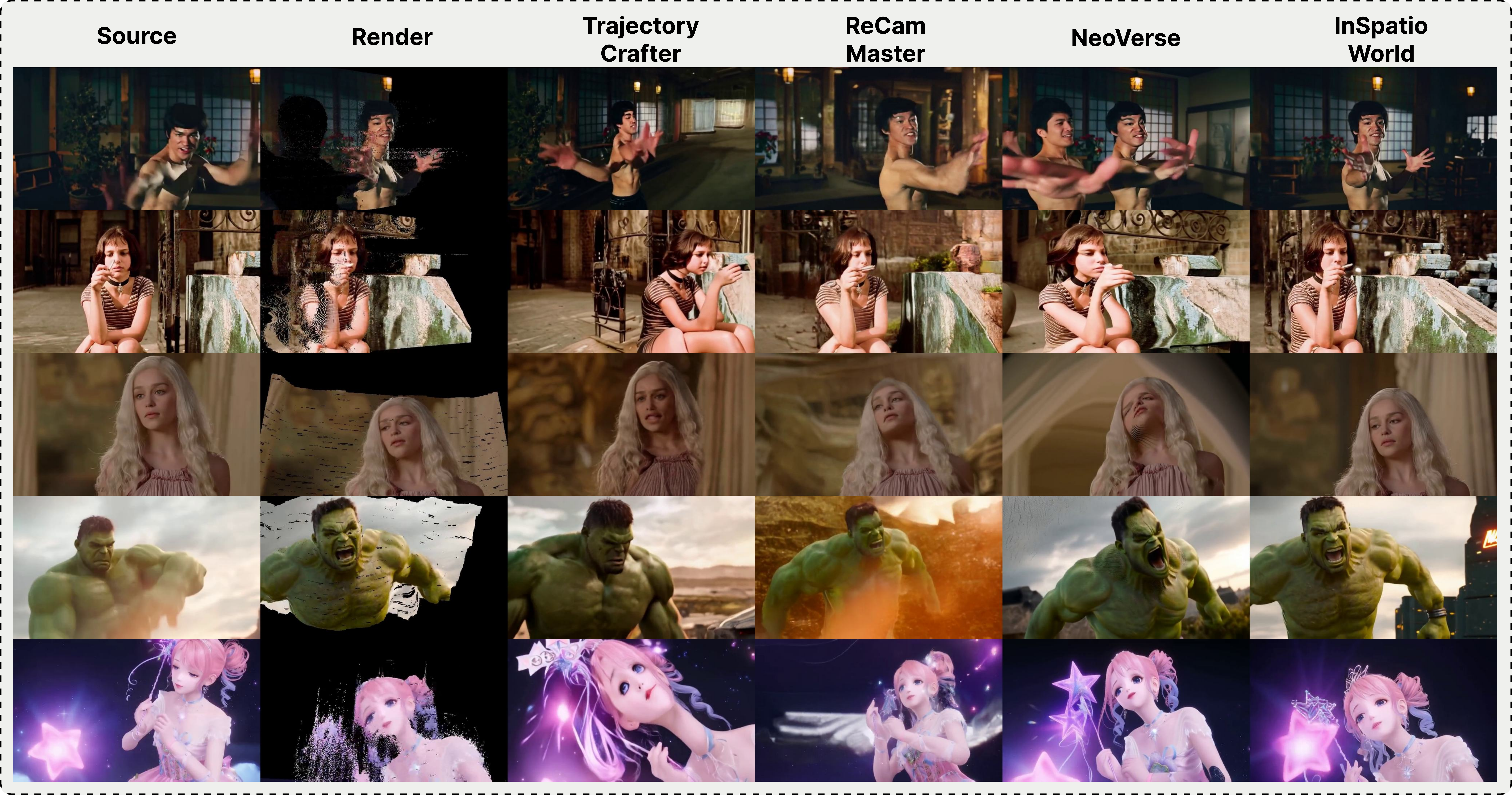}
    \caption{\textbf{Qualitative comparison on Camera Controlled Video Rerendering.} 
    Each row represents a distinct scene. From left to right: the first frame of the reference video, the warped final frame, and the final frames generated by TrajectoryCrafter, ReCamMaster, NeoVerse, and our method. 
    Compared to existing methods, our approach yields higher structural fidelity to the original scene and delivers significantly better textural details. 
    Simultaneously, it demonstrates superior instruction-following, achieving precise camera trajectories that are nearly identical to the rendered ground truth. The reference frames showcased are sampled from online video platforms and are utilized exclusively for academic demonstration purposes.
    }
    \label{fig:v2v_qualitative}
\end{figure}